\documentclass[sigconf]{acmart}

\setcopyright{none}
\renewcommand\footnotetextcopyrightpermission[1]{}
\makeatletter
\def\@copyrightspace{\relax}
\makeatother

\acmConference{}{}{}
\acmBooktitle{}
\acmYear{}
\copyrightyear{}
\acmISBN{}
\acmDOI{}

\AtBeginDocument{%
  }

\usepackage{multirow}
\begin{document}

\title{CAIM: Development and Evaluation of a Cognitive AI Memory Framework for Long-Term Interaction with Intelligent Agents}

\author{Rebecca Westhäußer}
\email{rebecca.westhaeusser@mercedes-benz.com}
\orcid{0009-0008-1282-0811}
\affiliation{
    \institution{Mercedes-Benz AG}
    \city{Böblingen}
    \country{Germany}
}

\author{Frederik Berenz}
\email{frederik.berenz@mercedes-benz.com}
\affiliation{
    \institution{Mercedes-Benz AG}
    \city{Böblingen}
    \country{Germany}
}

\author{Wolfgang Minker}
\email{wolfgang.minker@uni-ulm.de}
\orcid{0000-0003-4531-0662}
\affiliation{
    \institution{Ulm University}
    \city{Ulm}
    \country{Germany}
}

\author{Sebastian Zepf}
\email{sebastian.zepf@mercedes-benz.com}
\orcid{0000-0002-1268-146X}
\affiliation{
    \institution{Mercedes-Benz AG}
    \city{Böblingen}
    \country{Germany}
}

%\renewcommand{\shortauthors}{Westhäußer et al.}

%% The abstract is a short summary of the work to be presented in the article.
\begin{abstract}
Large language models (LLMs) have advanced the field of artificial intelligence (AI) and are a powerful enabler for interactive systems. However, they still face challenges in long-term interactions that require adaptation towards the user as well as contextual knowledge and understanding of the ever-changing environment. To overcome these challenges, holistic memory modeling is required to efficiently retrieve and store relevant information across interaction sessions for suitable responses. Cognitive AI, which aims to simulate the human thought process in a computerized model, highlights interesting aspects, such as thoughts, memory mechanisms, and decision-making, that can contribute towards improved memory modeling for LLMs.
Inspired by these cognitive AI principles, we propose our memory framework CAIM. CAIM consists of three modules: 1.) The Memory Controller as the central decision unit; 2.) the Memory Retrieval, which filters relevant data for interaction upon request; and 3.) the Post-Thinking, which maintains the memory storage.
We compare CAIM against existing approaches, focusing on metrics such as retrieval accuracy, response correctness, contextual coherence, and memory storage. The results demonstrate that CAIM outperforms baseline frameworks across different metrics, highlighting its context-awareness and potential to improve long-term human-AI interactions. 
\end{abstract}


\maketitle
% for arxiv submission
\makeatletter
\def\@headfoot{}%
\def\@oddhead{}%
\def\@evenhead{}%
\makeatother

\section{Introduction}
Advancements in LLMs, such as GPT-3.5 \cite{brown2020language}, are reshaping the interaction between humans and AI systems. These models allow direct and seamless communication through natural language. This offers new opportunities for human-computer interaction, contributing to their adoption across diverse domains \cite{chen2023gap, ning2024user, liu2023tim}. These advancements led to a growing area of research focused on utilizing LLMs to develop AI agents with human-like capabilities, such as decision-making and planning \cite{cui2024survey, zhong2024memorybank}.

While LLMs have advanced the field of AI, they still have shortcomings that hinder their ability to emulate human intelligence as they do not understand the ever-changing environment \cite{xie2023olagpt}. Furthermore, some LLMs are constrained by limited context windows, which negatively impacts their ability to understand and memorize historical information over extended periods \cite{xu2022long}, hindering their effectiveness in long-term interactions \cite{packer2023memgpt, sun2024can, zhong2024memorybank, liu2023tim, ram2023context}. Although models such as GPT-4o \cite{hurst2024gpt4} or Gemini \cite{team2024gemini} offer larger context windows, it is crucial to keep the input concise to ensure high performance, as longer inputs may reduce efficiency and potentially overwhelm the model \cite{lu2024insights}. Maintaining contextual understanding is critical to ensure meaningful interactions and understand user behavior \cite{zhong2024memorybank}. Therefore, remembering information from the past and correctly referencing it in ongoing conversations is crucial in long-term interactions \cite{bae2022keep}.

Without a long-term memory (LTM), LLMs also face challenges in delivering personalized responses. Personalization stands as an important connection that bridges the gap between humans and machines, as understanding the user's preferences is fundamental to adapting to their evolving needs and ensuring user satisfaction and engagement \cite{wozniak2024personalized, chen2024large, eapen2023personalization, wu2024understanding, kirmayr2025carmem}. Various studies emphasize the importance of personalization in building a relationship with AI assistants. For example, Lee et al. \cite{lee2012personalization} indicate that adding personalized services enhances user engagement with an assistant. Similarly, Ligthart et al. \cite{ligthart2022memory} and Kirmayr et al. \cite{kirmayr2025carmem} indicate that personalization fosters closeness, maintains the user's interest over time, and adds continuity to interactions across sessions. These challenges highlight the need for a context-aware memory mechanism that effectively manages input limits and proficiently draws on information from past conversations to improve long-term interactions and personalization.

In this context, methods such as reinforcement learning from human feedback \cite{wozniak2024personalized, wu2024understanding, salemi2024optimization} have shown promising approaches to improve the long-term capabilities of LLMs. Their practical application is often limited due to retraining the model, which is challenging and often impossible for models accessible only via API \cite{ram2023context, salemi2024optimization, lee2023prompted}. This paper focuses on retrieval-augmented methods as a practical and effective approach to equip LLMs with a memory mechanism without retraining the model, leaving the architecture unchanged. By utilizing external knowledge sources, memory-augmented methods address the limitation of a limited context window and enable the integration of past information into the current input prompt \cite{ram2023context, wang2023scm, liu2023tim, packer2023memgpt, sumers2023cognitive, xu2021beyond, lewis2020retrieval, zhang2023memory}.
While memory-augmented methods are already used to extend the general memory capabilities of LLMs, the existing approaches lack a context-aware mechanism and still fall short in achieving human-like performance in this regard. To address these limitations, insights from cognitive AI offer a promising perspective to enable LLMs to operate more contextually aware \cite{sun2024can, zhao2022emotion}. Cognitive AI intends to let computers think, reason, and make decisions similarly to humans, by mimicking their cognitive processes \cite{zhao2022emotion}. Therefore, incorporating insights from human cognition is promising to contribute to the development of more capable and reliable systems. Sun \cite{sun2024can} discusses the principles of Cognitive AI and their potential when bringing them together with LLMs. He argues that integrating these principles with LLMs can enhance several aspects, including information retention and retrieval, decision-making in dynamic environments, self-reflection, and interpersonal interactions.

Building on the potential of both memory-augmented methods and cognitive AI principles to achieve a more robust, human-like, and context-aware memory mechanism for LLMs, we propose CAIM, a \textbf{C}ognitive \textbf{AI} \textbf{M}emory Framework for LLMs. In addition to Cognitive AI, CAIM builds upon three existing memory-augmented approaches - MemoryBank, Think-in-Memory, and Self-Controlled Memory - by combining and extending their concepts to create a more holistic memory mechanism. Specifically, our approach consists of three modules (see Figure \ref{fig:workflow}) and addresses aspects of human memory, such as active decision-making, by positioning the LLM as a central decision unit involved in recalling memories, filtering retrieved information for contextual relevance, and determining which data to store in memory. Ultimately, CAIM aims to enhance long-term interactions by improving the memory capabilities of LLMs, which serves as a critical first step toward developing AI agents that act in a personalized manner and adapt to individual users over time.

% Challenges
In particular, CAIM addresses the following challenges of LLMs and existing approaches by combining a memory-augmented approach and cognitive AI principles.

\textbf{Challenge 1: Lack of an effective LTM.} Current LLMs are not designed to efficiently persist information across interactions, as a limited context window constrains them \cite{xu2022long}. This leads to forgetting user-specific information, preventing the model from delivering contextually relevant and personalized responses \cite{packer2023memgpt, zhong2024memorybank, liu2023tim}.

\textbf{Challenge 2: Response Correctness.} Existing memory-aug-mented approaches often struggle with consistent response correctness \cite{zhong2024memorybank, liu2023tim}. Without effective memory retrieval and prioritization mechanisms, irrelevant memories can be retrieved, decreasing the quality of responses.

% How CAIM solves these challenges
To address challenge 1, we designed an LTM mechanism to retain and recall information beyond a single interaction session, thereby contributing to ongoing efforts to create more intelligent and personalized AI assistants. We integrated a novel tagging system based on a predefined ontology to provide a more consistent categorization of memories and more accurate retrieval. To address challenge 2, we propose contextual and time-based relevance filtering to prioritize appropriate memories, improving the context awareness of CAIM. 
To support both challenges, we further integrate cognitive AI principles to mimic human memory processes, such as active decision-making and contextual retrieval. The main contributions of this paper can be summarized as follows.

% Contributions
\begin{itemize}
    \item We present a novel, more holistic memory mechanism based on an ontology, ensuring a more consistent tagging and retrieval process for more accurate responses.
    \item We introduce a contextual and time-based filtering mechanism that prioritizes relevant information, enhancing context awareness and enabling more effective memory retrieval and response generation.
    \item We propose a decision-making framework that enables LLMs to control memory usage, selecting and evaluating stored information based on cognitive AI principles.
    \item We conduct experiments with three LLMs on the public Generated Virtual Dataset that show that CAIM outperforms existing frameworks across different metrics such as response correctness and retrieval accuracy. Furthermore, we conducted ablation studies to assess component contributions of CAIMs Memory Controller and filtering mechanism.
\end{itemize}

% structure of paper
This work is structured as follows: First, we review relevant existing work in the fields of memory-augmented LLMs and cognitive AI principles. Second, we introduce CAIM's architecture. Third, we describe our experiments and present the insights we obtained. Finally, we discuss our findings, draw conclusions, and outline potential directions for future work.

\section{Related Work}
\subsection{Memory-augmented LLMs}
Numerous approaches have demonstrated that retrieval-augmented methods improved LLMs' long-term capability performance without the need for fine-tuning \cite{ram2023context, zhong2024memorybank, liu2023tim, wang2023scm, packer2023memgpt, xu2021beyond, ram2023context}. This is particularly useful since fine-tuning LLMs is complex, costly, and even impossible for LLMs accessible only via API \cite{ram2023context}. One example is MemoryBank \cite{zhong2024memorybank}, which introduced the differentiation of two memory types, long-term and short-term memory (STM), to resemble a more human-like memory mechanism for more natural interaction. Liu et al. \cite{liu2023tim} proposed their Think-in-Memory (TiM) that enables LLMs to remember and recall thoughts to incorporate them into the conversation without repeated reasoning. TiM consists of two key stages: recalling thoughts before generation and post-thinking after generation to update the memory. The self-controlled memory (SCM) proposed by Wang et al. \cite{wang2023scm} is a framework to enhance the ability of LLMs to recall only relevant information. Their approach incorporates a memory controller as a self-asking stage to determine whether memory retrieval is required based on the current user input. MemGPT, proposed by Packer et al. \cite{packer2023memgpt}, introduced a memory mechanism that utilizes different storage tiers to provide extended information. They employ strategies to prioritize context to improve long-term interactions. Xie et al. introduced OlaGPT, which simulates aspects of human cognition, including memory and the ability to learn from mistakes \cite{xie2023olagpt}. It aims to improve LLMs' problem-solving abilities by emulating human-like thought processes. OpenAI’s ChatGPT also incorporates memory functionality that allows the model to persist user-specific information across sessions, stored via an external memory layer \cite{OpenAI2024Memory}. Finally, Seo et al. \cite{seo2025prompt} proposed a prompt chaining framework designed to address the limitations of LLMs in maintaining context over extended periods. Their approach applies a multi-step reasoning process to enhance context-aware and personalized responses.

These approaches have demonstrated foundational work in improving memory capabilities, facing challenge 1. Still, they are limited in their ability to recall relevant memories, which negatively impacts response correctness. This can be problematic in long-term interactions, where evolving user preferences and shifting topics necessitate a more contextually aware memory management approach that utilizes context to retrieve relevant information \cite{du2024survey, chen2023gap, baldauf2007survey}. Furthermore, recent research highlights a need to enhance long-term capabilities in LLM agents further, for example, without repeated reasoning over the same historical data \cite{liu2023tim, pawar2024and}. CAIM aims to address these limitations of the existing memory-augmented approaches through a memory mechanism combining a tagging system and a context-aware retrieval process to improve long-term interactions, addressing challenge 2.

\subsection{Cognitive AI Principles}
Cognitive AI is derived from cognitive science and AI. Cognitive science is an interdisciplinary field that studies the processing and transmission of information within the human brain, addressing mental abilities of human beings such as language, perception, memory, attention, and reasoning \cite{chen2018cognitive, hwang2017big}. 
The overarching goal of cognitive AI is to simulate the human thought processes in a computerized model to learn, memorize, and respond to external changes to bring computing closer to human thinking \cite{zhao2022emotion, xie2023olagpt, hwang2017big}. To achieve this, recent research states that bridging AI with cognitive science is required to deepen the understanding of human cognition for more human-centric AI systems \cite{niu2024large, zhao2022emotion}. Niu et al. \cite{niu2024large} show that LLMs demonstrate abilities that resemble human cognition in various tasks, such as language processing and reasoning. Furthermore, they declare the potential of combining LLMs with cognitive architectures to create more robust, efficient, and adaptable AI systems.

According to Chen et al. \cite{chen2018cognitive}, the cognitive process of humans is divided into two stages: 1) becoming aware of external information as an input and 2) transmitting and processing the input to the brain for storage, analysis, and learning.
To be aware of the external information, a key principle of cognitive AI is a \textit{Decision Unit} that considers the current context to determine the next activity. Decision-making refers to the cognitive process of problem-solving activities to find the optimal solution \cite{zhao2022emotion, xie2023olagpt, sumers2023cognitive}. To transmit the information to the brain, the cognitive AI concept of a memory mechanism is essential \cite{hwang2017big}. The memory is often divided into several types of memories. Typical types are an \textit{STM}, called working memory in cognitive science \cite{baddeley2020working, atkinson1968human}, which reflects the current circumstances, and an \textit{LTM} that stores relevant knowledge for extended periods \cite{xie2023olagpt, xu2022long, atkinson1968human}. To access the stored information, the \textit{Retrieval Process} plays a crucial role as it allows humans to access contextually relevant past knowledge \cite{zhao2022emotion, xie2023olagpt}.
Another area of cognitive science is human learning \cite{hwang2017big}, leading to the cognitive AI principle Discovery, which involves implicit pattern recognition to find connections within the data and deepen the understanding of the user \cite{xie2023olagpt, zhao2022emotion}. Additionally, unlearning data, referring to memory forgetting, is crucial to minimize the use of outdated data to optimize the response correctness further \cite{sumers2023cognitive}. 

Based on these principles, cognitive AI needs representational structures for the artificial mind and computational procedures that operate on those structures \cite{hwang2017big}. CAIM combines a \textit{STM} and \textit{LTM} to enhance context-aware interactions and provide a representational structure. Furthermore, it integrates a \textit{Decision Unit} and a \textit{Retrieval Process} to select the most relevant memories based on the current context of the user input. Therefore, CAIM aims for a more holistic memory mechanism that aligns with cognitive AI principles by building upon human-like memory processes derived from cognitive science to improve overall performance, addressing challenges 1 and 2.

\section{CAIM}
\subsection{Architectural Foundation}
CAIM is based on a memory-augmented approach and follows the current trend of employing LLMs as the core of agent-based systems \cite{hatalis2023memory, li2024hello}. Additionally, we use in-context learning to adapt the LLM to tasks based on brief instructions via prompts \cite{lee2023prompted, brown2020language}. This approach enables the model to execute tasks without requiring retraining or modifications to its architecture, as data is added to the input \cite{lee2023prompted, ram2023context, xie2023olagpt, li2023practical}.

\subsection{Memory Structure}
The memory structure of CAIM is inspired by the common principle of cognitive science, which differentiates between \textit{LTM} and \textit{STM} \cite{atkinson1968human, sumers2023cognitive, hatalis2023memory, xu2022long}. This memory management addresses information storage outside the LLMs' context window, allowing the agent to retain and recall past information. Separate memories enable individual handling of historical data and ongoing sessions \cite{li2024hello, zhang2023memory, wang2023scm, hatalis2023memory, seo2025prompt}.

% STM & LTM
CAIM's \textit{STM} is dedicated to maintaining the context of the ongoing conversation \cite{atkinson1968human, baddeley2020working}. The STM temporarily stores recent conversation turns in a list, which is processed and transferred to the LTM by the Post-Thinking module. The \textit{LTM} retains historical data from past conversations over extended periods \cite{atkinson1968human}. It is inspired by the concept of inductive thoughts proposed by Liu et al. \cite{liu2023tim}. These thoughts are modeled after the human brain, which stores summarized thoughts of events instead of specific details \cite{zhong2024memorybank, liu2023tim}. Therefore, CAIM's LTM stores high-level event summaries from each conversation \cite{li2024hello, zhong2024memorybank}, ensuring that retrieved thoughts remain concise and do not exceed the limited context window of LLMs.

% LTM structure
The representational structure of our LTM organization consists of inductive thoughts, timestamps, and tags (see Table \ref{tab:memorystructure}). The timestamp enables the time-based prioritization of memories during retrieval. As a concept to organize the storage of key events, we introduce a tagging system with an LLM agent selecting contextually relevant tags from an ontology when storing new thoughts (see Appendix \ref{sec:Memory_Retrieval}, Prompt Select Tags). Due to the LLMs' powerful text-based processing capabilities, we decided on this approach instead of semantic similarity calculations, which is stated to be computationally intensive for long-term interactions \cite{liu2023tim, pawar2024and}. Findings from Li et al. \cite{li2024hello} further support our tag-based approach, indicating that topic-based retrieval mechanisms enhance retrieval accuracy.

\begin{table}[h!]
    \centering
    \caption{Memory Structure of CAIMs LTM}
    \begin{tabular}{ |p{2,3cm}|p{3,1cm}|p{1,9cm}| }
        \hline
        \textbf{Tags} & \textbf{Inductive Thought} & \textbf{Timestamp} \\
        \hline
        personal, identity, name & name is Emily & 2024-05-01 \\ 
        \hline
        movie, recommendations & System recommends 'Inception' & 2025-01-07 \\
        \hline
        food, preferences, likes & favorite food is pizza & 2025-04-09 \\
        \hline
        hobbies, piano & Emily enjoys playing piano & 2025-06-09 \\ 
        \hline
    \end{tabular}
    \Description{The table illustrates four example entries in CAIM’s long-term memory, each consisting of a set of tags, a stored inductive thought, and a timestamp. The tags categorize the memory content (e.g., personal, food, hobbies), enabling efficient retrieval and contextual relevance in user interactions.}
    \label{tab:memorystructure}
\end{table}

\begin{figure}[ht!]
    \centering
    \includegraphics[width=0.45\linewidth]{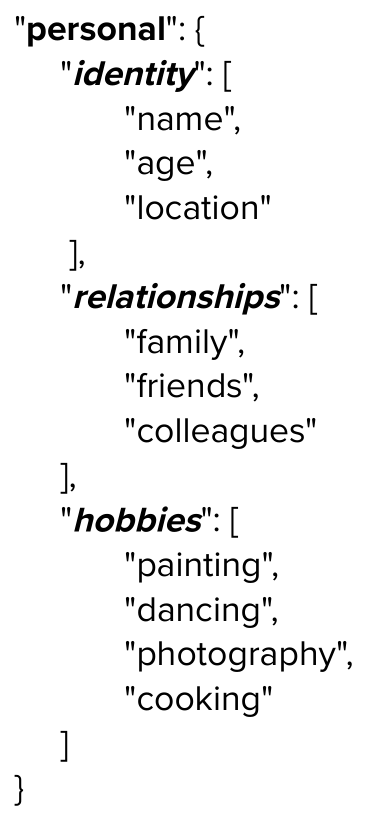}
    \caption{Example of CAIM's ontology structure}
    \Description{The figure shows a structured excerpt of CAIM’s ontology, illustrating three depth levels: category, subcategory, and attributes. It highlights how user-related concepts are semantically organized to support memory tagging and contextual retrieval.}
    \label{fig:ontology}
\end{figure}

% Ontology
The ontology used for CAIM's tagging system is generated by an LLM and structured in JSON format. We instructed the model to develop a compact yet comprehensive ontology suitable for a chatbot that allows users to discuss various topics. Additionally, the ontology was designed with a hierarchical structure limited to three depth levels and restricted to single-word terms, which we refer to as tags. The three depth levels are: categories (e.g., personal, technology, food), which are divided into subcategories (e.g., identity, relationships, hobbies), and attributes (e.g., name, age, location) to refine the context further. Figure \ref{fig:ontology} shows an exemplary excerpt of the ontology. The ontology aims to reduce the variability of tags and ensures that the classification of memories remains more consistent. This supports the structured knowledge representation and retrieval \cite{guarino2009ontology}, allowing LLM agents to categorize and relate memories to broader contexts. 

A depth of three levels was chosen to limit the complexity, as the ontology serves as an initial structure for organizing user input. However, the ontology is extensible, allowing the LLM agent to add new single-word tags when the agent determines it is necessary to describe new thoughts. To determine whether an update of the ontology is required, the LLM agent is instructed via prompt (see Appendix \ref{sec:Memory_Retrieval}, Prompt Ontology Expansion) to analyze the user input to determine whether the existing tags in the ontology sufficiently describe the input. If they do, the ontology remains unchanged, otherwise, the model dynamically extends it by adding new tags that the LLM agent generates based on the current user input. This functionality minimizes the risk of incorrectly classifying an inductive thought or providing an insufficient number of tags.

\subsection{Modules}
\begin{figure*}[ht!]
    \centering
    \includegraphics[width=1\linewidth]{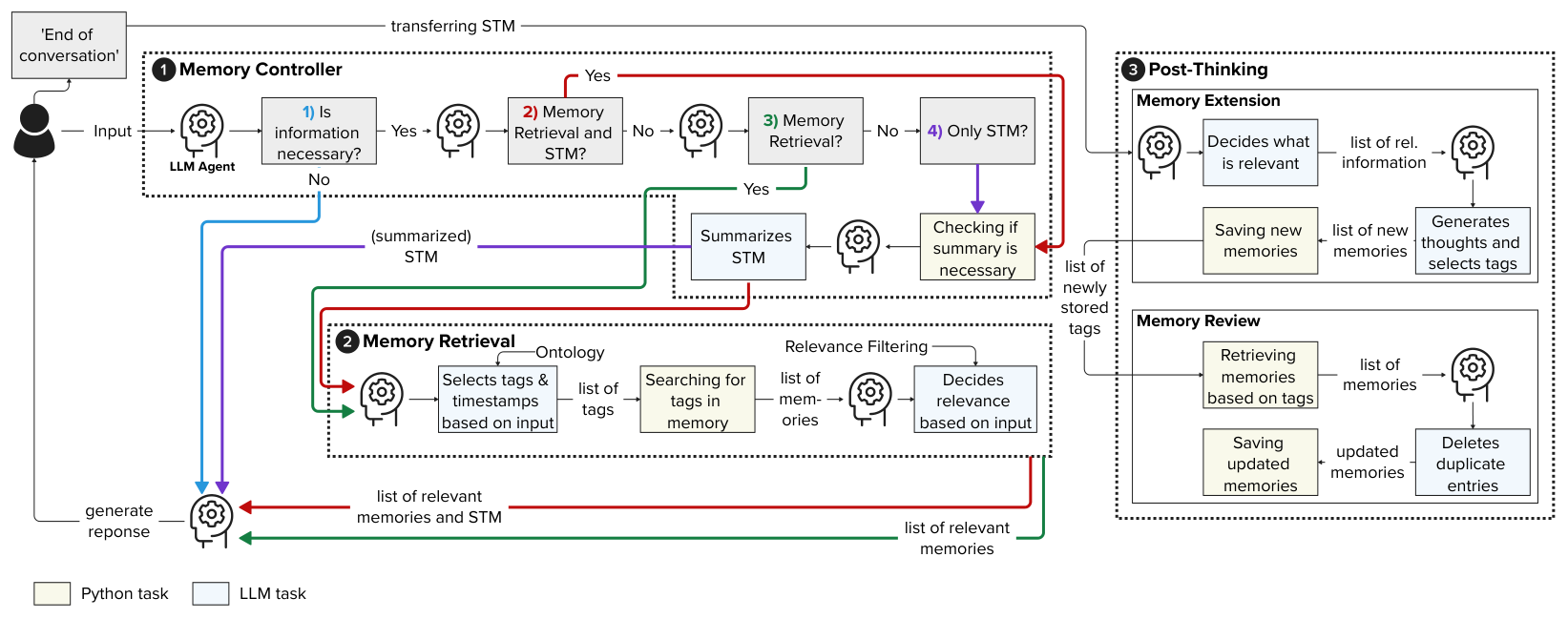}
    \caption{Workflow of CAIM - Illustration of the three modules, with colors highlighting the different possible paths within the system.}
    \Description{CAIM architecture contains three modules: Memory Controller, Memory Retrieval, and Post-Thinking to enable a context-aware memory mechanism.
    Input is managed by the Memory Controller, which determines whether memory retrieval or summarization of short-term memory is needed, and orchestrates flow to downstream modules. The Memory Retrieval module uses ontology-based tag selection and relevance filtering to return relevant memories. Post-Thinking includes Memory Extension, which saves new and relevant information, and Memory Review, which updates or removes duplicate memories. Outputs from all modules support response generation by the LLM agent.}
    \label{fig:workflow}
\end{figure*}

The concept of CAIM is based on existing frameworks, such as MemoryBank \cite{zhong2024memorybank}, TiM \cite{liu2023tim}, and SCM \cite{wang2023scm}, integrating key aspects of these approaches. As illustrated in Figure \ref{fig:workflow}, CAIM is structured into three modules designed to work together to address the outlined challenges, including the lack of LTM in LLMs and the need for more accurate and contextually relevant responses.

\subsubsection{\textbf{Memory Controller}}
The Memory Controller is a self-asking mechanism inspired by SCM \cite{wang2023scm}. This module allows the LLM agent to determine when to retrieve memories, resembling the \textit{Decision Unit} principle of cognitive AI. 
As shown in Figure \ref{fig:workflow}, the interaction with CAIM starts with a user input, which is transferred to the first module, the Memory Controller, starting with an LLM agent that is instructed via prompts to evaluate the user input and determine whether additional information is needed to ensure a contextually appropriate response (see Appendix \ref{sec:Memory_Controller}). Depending on the specific prompt, the agent decides between different options: For example, the prompt might ask whether the response requires recalling historical information from earlier in the conversation, or if the current context alone is sufficient. The agent always selects between two options provided in the prompt. Based on these evaluations, the agent selects one of the following options:

\begin{enumerate}
  \item Is additional information necessary?
  \item Are both historical information (LTM) and conversation context (STM) necessary?
  \item Is only historical information necessary?
  \item Is only conversation context necessary?
\end{enumerate}

For the first option, an agent is instructed to determine whether previous information is required to provide a relevant response or if the user request can be addressed without relying on prior context (see Appendix \ref{sec:Memory_Controller}, Prompt Information necessary). If no prior context is necessary, the agent generates a response using its internal knowledge base. For example, when asked ``What is the content of the movie Inception?'' the model does not require additional context and can rely on its pretrained knowledge. If additional information is required, such as in the query ``What movie did I watch on April 2nd?''. In that case, an agent gets prompted with a second instruction to evaluate whether both historical information and the ongoing conversation context are needed to generate an appropriate response (see Appendix \ref{sec:Memory_Controller}, Prompt Retrieval and Conversation). If both combined are not required to address the user’s request, a third prompt is used to determine whether the user input references a prior event, which only requires historical data to respond correctly (see Appendix \ref{sec:Memory_Controller}, Prompt Retrieval). If none of these scenarios apply, an LLM agent uses the STM, which contains the context of the ongoing conversation, to address the user input. The Memory Controller also determines whether to process the full STM content or a summarized version, as LLMs have a limited input length \cite{wang2023scm, liu2023tim}.

% why important?
Overall, the Memory Controller plays a crucial role in selective memory retrieval, ensuring that memories are retrieved only when needed and relevant. This helps to reduce the amount of data and prevent decreased performance caused by large input \cite{wang2023scm}.

\subsubsection{\textbf{Memory Retrieval}}
The Memory Retrieval is responsible for reading information from an external LTM to extend the contextual knowledge of the LLM agent \cite{sumers2023cognitive, zhang2023memory}, resembling the \textit{Retrieval Process} principle of cognitive AI. We integrated filtering mechanisms to prioritize only contextually and time-based relevant data regarding the user input to ensure the generation of a coherent response, addressing challenge 2.

As illustrated in Figure \ref{fig:workflow} (2. Memory Retrieval), the LLM agent is prompted to analyze the user input and select contextually relevant tags from the ontology. To achieve this, the agent receives both the user input and the ontology within its prompt and is instructed to identify tags that summarize the main topic or intent of the message, enabling a context-aware retrieval process (see Appendix \ref{sec:Memory_Retrieval}, Prompt Select Tags). 
Furthermore, another agent is instructed to determine whether the input includes a time-based unit (e.g., ``yesterday'', ``June 9th''). If such references are present, the LLM agent generates corresponding timestamps. A Python function uses these tags and timestamps to retrieve memories from the LTM that share identical tags or timestamps.
The following agent obtains both the user input and the retrieved memories, determines which memories are contextually relevant, and returns only those that align with the user query (see Appendix \ref{sec:Memory_Controller}, Prompt Relevance of Memories). Lastly, the list of relevant memories is passed to the final agent to generate a response.

% why important?
In summary, the Memory Retrieval mechanism ensures that only contextually and time-based relevant data is forwarded for response generation, which is essential for improving response correctness, addressing challenge 2.

\subsubsection{\textbf{Post-Thinking}}
Maintaining an up-to-date memory in long-term conversations is essential for human-like dialogues \cite{bae2022keep, park2023generative}. To address this, we integrated a Post-Thinking module inspired by TiM \cite{liu2023tim} to maintain the LTM with operations such as insert and merge \cite{bae2022keep}. These operations are used to store new data and merge duplicate entries. The Post-Thinking module is structured into two phases: Memory Extension and Memory Review.

As illustrated in Figure \ref{fig:workflow} (3. Post-Thinking), the module starts with the Memory Extension process, which stores new memories in the LTM. The process is initiated at the end of a conversation when the STM is transferred to an LLM agent to determine which elements are relevant for future usage. This agent is instructed to delete casual exchanges, such as greetings and small talk, focusing only on extracting key events of personal information as concise sentences (see Appendix \ref{sec:Post_Thinking}, Prompt Key Events). The resulting list of key events is passed to the following LLM agent, which generates inductive thoughts and selects up to three tags from the ontology. The inductive thought provides a more concise representation of the key events, thereby preventing the memory from becoming overloaded with detailed information.
The LLM agent is instructed to process each key event individually to generate inductive thoughts. First, it selects up to three tags from the provided ontology that summarize the main content of the key event, following the same approach used in the Memory Retrieval process. Second, it generates a concise summary of the key event, referred to as the inductive thought. Third, the agent uses the exact timestamp associated with the key event. Lastly, the agent formats the output according to CAIM’s memory structure, as illustrated in Table \ref{tab:memorystructure}. Finally, the memories are further processed by a Python function and stored in the LTM.

Following the Memory Extension, the following process is the Memory Review, which determines which information to keep in the LTM. This process first retrieves all memories associated with one of the tags assigned to the newly stored data with a Python function. These memories are passed to an LLM agent, which is instructed to identify memories with the same meaning and merge these duplicate entries. The agent is prompted to format the output according to CAIM’s memory structure. Finally, a Python function processes and stores the updated memories in the LTM.

% why important?
As stated by Packer et al. \cite{packer2023memgpt}, consistency is crucial in memory management since new facts, preferences, and other information should align with previously stored data, indicating the importance of a review process. The Post-Thinking module maintains a more natural memory management process by storing only relevant data, leading to an up-to-date memory \cite{liu2023tim}. Furthermore, this mechanism prevents unnecessary growth of the LTM, which could otherwise slow down the Memory Retrieval and overall system performance.

% summary
Approaching a more holistic memory modeling, these three modules form the core memory mechanism of CAIM, combining and extending mechanisms from existing work as well as adapting functional concepts towards the main capabilities of LLMs. This enables both the summarization and storage of relevant information from past conversations and context-aware filtering and retrieval of relevant information during the current conversation.

\section{Experiment}
\subsection{Experimental Setup}
\subsubsection{\textbf{Dataset}}
To evaluate CAIM, we used the public Generated Virtual Dataset (GVD), introduced by Zhong et al. \cite{zhong2024memorybank}. The dataset contains long-term conversation data involving 15 users interacting with a virtual assistant over ten days. The test set includes 100 English questions that refer to content from the conversation. We chose this dataset to compare CAIMs performance in recalling memories and generating contextually appropriate answers against MemoryBank and TiM - as CAIM builds upon these frameworks, and both have been evaluated on this dataset. Although CAIM also builds upon SCM, a direct comparison with SCM was not performed in this evaluation due to differences in the evaluation metrics and datasets used, which are not directly comparable to those of MemoryBank and TiM. Therefore, we focused on MemoryBank and TiM to ensure a consistent comparison, as both use the same dataset and evaluation metrics.

\subsubsection{\textbf{LLMs}}
For the evaluation, we integrated the following three LLMs into our framework to demonstrate the effectiveness of CAIM across different models:
\begin{itemize}
    \item \textbf{GPT} \cite{brown2020language}, a language model developed by OpenAI, based on the Transformer architecture. The model is designed for multilingual usage, text understanding and generation. We use \textbf{GPT-3.5 turbo} to compare CAIM against existing approaches and \textbf{GPT-4o} to show CAIM's performance with the latest version.
    \item \textbf{ChatGLM} \cite{glm2024chatglm}, which is an open-source language model based on the General Language Model (GLM) framework. It is primarily used for Chinese and English language tasks.
\end{itemize}

\subsubsection{\textbf{Metrics}}
Following MemoryBank \cite{zhong2024memorybank} and TiM \cite{liu2023tim}, we adopted three metrics to evaluate the performance of our proposed framework. Additionally, we introduce a fourth metric to assess the quality of memory storage and investigate the relationship between memory storage and retrieval. The final set of metrics is as follows.
\begin{itemize}
    \item \textbf{Retrieval Accuracy} evaluates whether the relevant memories are successfully retrieved (labels: 0 = no; 1 = yes).
    \item \textbf{Response Correctness} evaluates whether the response contains the correct answer to the question (labels: 0 = wrong; 0.5 = partial; 1 = correct). 
    \item \textbf{Contextual Coherence} evaluates whether the response is naturally and coherently structured (labels: 0 = not coherent; 0.5 = partially coherent; 1 = coherent).
    \item \textbf{Memory Storage} evaluates whether the relevant information is stored inside the LTM (labels: 0 = not stored; 0.5 = partially stored; 1 = stored)
\end{itemize}

\subsubsection{\textbf{Evaluation}}
The evaluation follows the procedure of MemoryBank and TiM \cite{zhong2024memorybank, liu2023tim} and is designed as a fact-checking task. Three human annotators with expertise and experience in LLM application and research independently rated the retrieved memories, responses, and memory storage across different models for each question to evaluate the performance. Before the evaluation, the annotators were provided with detailed instructions, including definitions of evaluation metrics and example cases to ensure consistency. 

In cases where all annotators assigned identical ratings, the rating was directly accepted. In cases of disagreement among annotators, the differences were analyzed to identify their underlying causes. The annotators discussed their perspectives, referring to the evaluation metrics and example cases provided during the training phase. If a consensus could not be reached, the final decision was made based on majority voting. This process ensured that disagreements were systematically addressed while maintaining the reliability and objectivity of the evaluation.

To measure inter-rater agreement and the reliability of our annotations, we calculated Percent Agreement and Intra-class Correlation Coefficient (ICC) \cite{gisev2013interrater, chaturvedi2015evaluation} across all evaluation metrics and models. The Percent Agreement provides an overall measure of how often the annotators agreed on the classification of each evaluation metric \cite{gisev2013interrater, chaturvedi2015evaluation}. Following the flow chart for selecting an appropriate ICC by McGraw et al. \cite{mcgraw1996forming}, we used the two-way fixed effect model for ICC, where all raters assess each item.

Table \ref{tab:inter_annotator_agreement} presents the values of the inter-rater agreement metrics for each model, calculated across the four evaluation metrics: Retrieval Accuracy, Response Correctness, Contextual Coherence, and Memory Storage.

\begin{table*}[h!]
    \centering
    \caption{Inter-rater Agreement Scores across four Evaluation Metrics for each LLM}
    \begin{tabular}{|c|c|c|c|c|c|c|}
        \hline
        {} & \multicolumn{2}{|c|}{GLM} & \multicolumn{2}{|c|}{GPT-3.5 turbo} & \multicolumn{2}{|c|}{GPT-4o} \\
        \hline
        {} & \textbf{Percent Agr.} & \textbf{ICC} & \textbf{Percent Agr.} & \textbf{ICC} & \textbf{Percent Agr.} & \textbf{ICC}\\
        \hline
        \textbf{Memory Retrieval} & 98\% & 96.9\% & 96\% & 90.5\% & 93\% & 78.1\%\\
        \hline
        \textbf{Response Correctness} & 89\% & 94\% & 88\% & 91.3\% & 84\% & 84.2\%\\
        \hline
        \textbf{Contextual Coherence} & 64\% & 50.9\% & 93\% & 5.8\% & 97\% & -0.6\%\\
        \hline
        \textbf{Memory Storage} & 86\% & 89.1\% & 83\% & 67.9\% & 76\% & 53.2\%\\
        \hline
    \end{tabular}
    \Description{The table reports inter-rater agreement for GLM, GPT-3.5 Turbo, and GPT-4o using Percent Agreement and Intraclass Correlation Coefficient (ICC) across the metrics Memory Retrieval, Response Correctness, Contextual Coherence, and Memory Storage.}
    \label{tab:inter_annotator_agreement}
\end{table*}

% Memory Retrieval
For the Memory Retrieval, the Percent Agreement ranged from 93\% for GPT-4o to 96\% for GPT-3.5 turbo and 98\% for GLM. For this metric, annotators had two choices (0 or 1), indicating whether the correct memory was retrieved. The ICC values, which measures absolute agreement between raters \cite{mcgraw1996forming}, range from 78.1\% for GPT-4o to 90.5\% for GPT-3.5 turbo and 96.9\% for GLM.

% Response Correctness
Response Correctness also demonstrates high levels of agreement for both metrics. The Percent Agreement values ranged from 84\% for GPT-4o to 89\% for GLM, reflecting a strong alignment among annotators. The ICC values, which ranged from 84\% (GPT-4o) to 94\% (GLM), further underscore the reliability of these ratings.

% Memory Storage
Memory Storage showed Percent Agreement values of 76\% (GPT-4o), 83\% (GPT-3.5 turbo), and 86\% (GLM). The ICC values ranged from 53\% (GPT-4o) to 89\% (GLM), indicating greater variability in annotator judgments compared to the other metrics.

% Contextual Coherence
A notable difference in Percent Agreement occurs for the Contextual Coherence metric for GLM, where agreement drops to 64\% due to inconsistencies in the model-generated responses, leading to a lower agreement rate. However, the values for the GPT models remain high with 93\% for GPT-3.5 turbo and 97\% for GPT-4o. 
The ICC values for Contextual Coherence reveal a limitation of the metric in cases of low rating variability. GLM achieves an ICC of 50.9\%, suggesting a moderate level of agreement \cite{gisev2013interrater, chaturvedi2015evaluation}. In contrast, GPT-3.5 turbo demonstrates a low ICC of 5.8\%, indicating poor reliability, while GPT-4o shows an ICC of -0.6\%. However, these ICC values are misleading, as the high Percent Agreement among annotators demonstrates that they mostly agreed. According to Bajpai et al.\cite{chaturvedi2015evaluation}, low ICC values suggest that the metric fails to reflect agreement effectively when rating variability is low. Following these findings, a detailed examination of our data reveals that, for example, annotators agreed on 93 out of 100 instances for the Contextual Coherence of GPT-3.5 turbo, resulting in a 93\% Percent Agreement. They consistently assigned the highest score (1), reflecting a strong contextual coherence in the model’s responses. This high level of agreement indicates a reliable and objective assessment, but leads to a low ICC value (5.8\%) due to the limited variability in the ratings.

Overall, the Percent Agreement and ICC show a strong level of inter-rater agreement across all models and evaluation metrics, demonstrating the reliability of the annotation process.

\subsection{Results}

\begin{table*}[h!]
    \centering
    \caption{Comparison of CAIM and Baseline Models on GVD}
    \begin{tabular}{ |c|c|c|c|c|c| } 
        \hline
        \textbf{Model} & \textbf{Memory} & \textbf{Retrieval Accuracy} &\textbf{ Response Correctness} &\textbf{ Contextual Coherence} & \textbf{Memory Storage} \\
        \hline
        \multirow{3}{2cm}{ChatGLM} & MemoryBank & 80.9\% & 43.8\% & 68\% & - \\ 
         & TiM & 82\% & 45\% & 73.5\% & - \\ 
         & CAIM & \textbf{67.6\%} & \textbf{60\%} & \textbf{86.5\%} & 79\% \\ 
        \hline
        \multirow{2}{2cm}{GPT-3.5 turbo} & MemoryBank & 76.3\% & 71.6\% & 91.2\% & - \\ 
         & CAIM & \textbf{83.3\%} & \textbf{81.3\%} & \textbf{98.3\%} & 85.2\% \\ 
        \hline
        GPT-4o & CAIM & 88.7\% & 87.5\% & 99.5\% & 90\% \\ 
        \hline
    \end{tabular}
    \Description{The table presents evaluation results for CAIM, MemoryBank, and TiM on the GVD benchmark using ChatGLM, GPT-3.5 turbo, and GPT-4o. Metrics include Retrieval Accuracy, Response Correctness, Contextual Coherence, and Memory Storage. Notably, CAIM achieves a response correctness of 60\% with ChatGLM, compared to 43.8\% and 45\% for MemoryBank and TiM, respectively. For GPT-3.5 turbo and GPT-4o, CAIM yields the highest scores across all reported metrics. Memory Storage is only reported for CAIM, reflecting its ability to persist information in long-term memory.}
    \label{tab:comparison_results}
\end{table*}

\subsubsection{\textbf{Comparison Results}}
The comparison results of CAIM and baseline models on GVD are presented in Table \ref{tab:comparison_results}. The comparison shows that CAIM surpasses the baseline models across different metrics, demonstrating CAIM's effectiveness.

% memory storage
\textbf{Memory Storage:} The results show that CAIM's modules enable the underlying model to retain information successfully across multiple sessions. This capability is demonstrated by the memory storage metric, which shows that 79\% (GLM), 85.2\% (GPT-3.5 turbo), and 90\% (GPT-4o) of relevant conversation information is successfully retained. These results further indicate that the GPT models are superior to GLM. 

% retrieval accuracy + limitation CAIM + GLM
\textbf{Retrieval Accuracy:} CAIM achieves a retrieval accuracy of 83,3\% with GPT-3.5 turbo, surpassing MemoryBank's 76.3\%. GPT-4o achieves an even higher retrieval accuracy of 88.7\%.
GLM reveals a limitation of CAIMs memory retrieval mechanism, showing a retrieval accuracy of 67.6\%, which is lower than for TiM (82\%) and MemoryBank (80.9\%). Notably, relevant memories were present in the LTM for 79\% of the questions. The performance gap with GLM is mainly caused by the fact that GLM struggles to consistently select tags from the ontology and respond consistently in predefined formats (e.g. ``tag1, tag2, tag3''). GLM often generates tags, including instances where tags are generated in Chinese despite the probing question being in English. These inconsistencies lead to mismatches between tags during the storage and retrieval of memories. Since the consistent selection of predefined tags is essential for the effectiveness of our proposed memory mechanism, these issues hinder CAIM's performance with GLM.
In summary, these results demonstrate that CAIMs modules enable the successful retrieval of user-specific context across multiple sessions. The retrieval accuracy indicates that the information is likely to be retrieved once stored in the LTM. This capability forms the basis for generating context-aware responses tailored to individual users.

% response correctness
\textbf{Response Correctness:} Table \ref{tab:comparison_results} shows that the responses of CAIM achieve a correctness rate of 60\% for GLM, representing an improvement of over 15\% compared to TiM (45\%) and MemoryBank (43.8\%), although the retrieval accuracy was worse (67.6\%). For GPT-3.5 turbo, CAIM achieves a response correctness of 81.3\%, increasing this metric by nearly 10\% compared to MemoryBank (71.6\%). Furthermore, CAIM achieves its highest response correctness of 87.5\% with GPT-4o. These numbers demonstrate the effectiveness of the integrated filtering mechanisms during memory retrieval, which prioritizes relevant memories based on context and time. These results demonstrate that CAIM improves response correctness across all models, successfully addressing challenge 2.

% contextual coherence
\textbf{Contextual Coherence:} CAIM improves the contextual coherence across all underlying models. For GLM, our approach achieves a coherence of 86.5\%, marking an improvement compared to TiM (73.5\%) and MemoryBank (68\%). GPT-3.5 turbo further increases the contextual coherence of CAIM up to 98.3\%, surpassing MemoryBank's 91.2\%. These results indicate that CAIM's retrieval module and filtering mechanism enable the LLM to provide a contextually relevant reply using the retrieved data. This demonstrates that with CAIM, the interactions are more naturally embedded into the conversation context, aligning with the goals of cognitive AI.

% Summary
These comparison results across different models and metrics demonstrate that CAIM has a robust long-term mechanism, effectively addressing challenge 1.

% Analyze number of words 
As additional analysis, we considered the number of words and entries for the memory with each model to gain deeper insights into their impact on the memory composition and amount.

\begin{table}[h!]
    \centering
    \caption{Storage Entries and Word Counts of CAIMs LTM across Models}
    \begin{tabular}{ |p{2cm}|p{2,7cm}|p{2,4cm}| }
        \hline
        \textbf{Model} & \textbf{Number of entries} & \textbf{Number of words} \\
        \hline
        ChatGLM & 330 & 4220 \\
        \hline
        GPT-3.5 turbo & 344 & 4757 \\
        \hline
        GPT-4o & 599 & 4230 \\ 
        \hline
    \end{tabular}
    \Description{The table presents the number of storage entries and total word counts in CAIM’s long-term memory for each language model. GPT-4o generates the highest number of entries (599), while GPT-3.5 turbo produces the largest word count (4757). This illustrates differences in memory-writing behavior across models despite using the same memory architecture and experimental setup.}
    \label{tab:word_count}
\end{table}

% GPT-4o
Table \ref{tab:word_count} presents the total number of entries and words, as well as the average number of words per entry for each model. GPT-4o generated the highest number of entries (599) across the 15 users despite having almost the lowest total word count (4230). GPT-4o mainly stores concise, inductive thoughts rather than detailed sentences, such as GLM. 
Compared to the other models, GPT-4o achieves the highest performance across all metrics in Table \ref{tab:comparison_results}, which highlights the model's efficiency in storing and retrieving relevant information, focusing mainly on quality (relevant, concise thoughts) over quantity (detailed sentences). This indicates that a larger number of entries with more concise thoughts is beneficial for long-term interaction, enabling the model to generate relevant, context-aware responses over extended periods.

% GPT-3.5 turbo
GPT-3.5 turbo demonstrates the highest word count (4757) with 344 entries. Compared to the performance metrics presented in Table \ref{tab:comparison_results}, GPT-3.5 turbo shows slightly lower performance than GPT-4o. This might indicate that storing more data does not necessarily lead to better performance and may instead indicate inefficiencies in memory storage and retrieval due to the storage of irrelevant data.

% GLM
GLM has the fewest entries (330) and the lowest total word count (4220). Compared to the GPT models, GLM primarily stores summarized sentences of the conversation rather than concise inductive thoughts, leading to a lower breadth of information. 
This lack of information negatively impacts performance across all metrics (see Table \ref{tab:comparison_results}). Furthermore, GLM struggles to manage split data and connect contextually related thoughts, highlighting a limitation of the underlying model rather than CAIM itself.

\subsubsection{\textbf{Ablation Study}}
We conducted an ablation study to evaluate the contribution of the Memory Controller and the Relevance Filtering component of CAIM, following the same annotation procedure as in the main experiment. We selectively removed each component to evaluate the performance on the GVD, using GPT-4o as the underlying LLM, as it yielded the highest performance across all evaluation metrics when integrated with CAIM. The Post-Thinking module remained unchanged across all system configurations to ensure that the same LTM data was used for each user as in the main experiment. This decision enables a controlled comparison between configurations by maintaining consistent memory content across all ablations. Table \ref{tab:ablation} presents the results of the ablation study for the evaluation metrics: Retrieval Accuracy, Response Correctness, and Contextual Coherence.

\begin{table}[h!]
    \centering
    \caption{Results of Ablation Study for GPT-4o}
    \begin{tabular}{ |p{2,3cm}|p{1,4cm}|p{1,7cm}|p{1,7cm}| }
        \hline
        \textbf{System Configuration} & \textbf{Retrieval Accuracy} & \textbf{Response Correctness} & \textbf{Contextual Coherence} \\
        \hline
        Full System & 88.7\% & 87.5\% & 99.5\% \\
        \hline
        w/o Memory Controller & 85.7\% & 78.2\% & 90.3\% \\
        \hline
        w/o Relevance Filter & 64.3\% & 63.5\% & 90.2\% \\ 
        \hline
    \end{tabular}
    \Description{The table shows ablation results across three evaluation metrics, comparing the full CAIM system with two configurations: one without the Memory Controller and one without the Relevance Filter. The findings highlight the individual contributions of each component to Retrieval Accuracy, Response Correctness, and Contextual Coherence.}
    \label{tab:ablation}
\end{table}

% w/o Memory Controller
Without the Memory Controller, CAIM consistently has access to the conversation history and always performs a Memory Retrieval, resulting in a similar Retrieval Accuracy of 85.5\% compared to the full system with 87.5\%. The Response Correctness drops from 87.5\% to 78.2\% when not involving the Memory Controller, which indicates that careful consideration whether additional information is needed or not is crucial for reliably obtaining proper responses. The Contextual Coherence remains high with 90.3\%, indicating that the underlying LLM can still generate coherent responses despite increased amount of data. Overall, these insights highlight the positive impact of the Memory Controller, which regulates the amount of information passed to the LLM and thereby helps to avoid overload.

% w/o Relevance Filtering
Removing the filtering mechanism causes a drop of the Retrieval Accuracy from 88.7\% to 64.3\%. Further, the Response Correctness is reduced from 87.5\% to 63.5\%. These results indicate that removing the relevance- and time-based retrieving and filtering degrades the Response Correctness. The Contextual Coherence remains high at 90.2\%, indicating that the LLM can still generate coherent responses, even when missing relevant knowledge. These results indicate that Relevance Filtering is crucial for identifying relevant information to generate a context-aware response.

\subsubsection{\textbf{Qualitative Findings}}
In addition to the findings from our numeric analysis for the considered metrics, we examined the unsuccessful replies across the different models in more detail. In this section, we report the respective findings, which help to better understand the strengths and weaknesses of particular models and potential areas for further improvement of CAIM.

% rhetorical questions
\textbf{Rhetorical Questions:} The dataset contains six rhetorical, negative tag questions to test contextual understanding. Examples of such questions are: 'I don't like working out, right?' (Sunny, Q. 26) or 'May 4th is not my birthday, right?' (Jack, Q. 77). Our findings reveal different behaviors among the models for these questions. For instance, GPT-4o contradicts the user by refuting their statement and responding correctly, such as in the following example: 'Actually, you enjoy working out! [...]' (Sunny, A. 26). GPT-3.5 turbo partially refutes the user's statement, achieving partially correct responses. In contrast, GLM failed to provide accurate responses as the model always agreed with the user, often replying with statements such as 'May 4th is not your birthday, correct.' (Jack, A. 77) or 'Yes, that's correct. I understand that you don't like warm weather.' (John Zhang, A. 43).
CAIM successfully stored the relevant information related to the rhetorical questions in 17 out of 18 cases and retrieved the stored memories correctly in 12 out of 18 cases. As previously mentioned, most incorrect retrievals were associated with GLM (5 out of 6), which struggled with selecting predefined tags from the ontology. This suggests that the performance issue stems from the limited capabilities of the underlying LLM.

% questions without data
\textbf{Questions without relevant data:} The GVD also contains two questions without relevant information in the provided conversational data to test how CAIM handles the absence of knowledge. These questions are: 'What dish did I make on May 5th?' (John Zhang, Q. 47) and 'On April 28th, did I share with you my experience of making twice-cooked pork?' (Gary, Q. 64).
Our analysis reveals different behaviors among the models. GPT-4o generates one hallucinated response: 'On May 5th, you made chicken enchiladas.' (John Zhang, A. 47), indicating the model's inability to acknowledge its lack of knowledge. In contrast, GPT-3.5 turbo demonstrates an awareness of the absence of relevant data, providing correct responses like 'I'm sorry, but I don't have a record of us discussing twice-cooked pork on April 28th. [...]' (Gary, A. 64). Similarly, GLM acknowledges its limitations, aligning with GPT-3.5 turbo.

% Problems
\textbf{Detailed Answers:} The dataset also contains questions that require detailed answers, such as the summarization of movies and books like 'I shared with you a literary film [...], what is its content?' (Emily, Q. 5), and provide an exact list of recipes or recommended items, such as '[...] Do you remember the specific recipe?' (Ivy, Q. 53) or 'What methods have you told me about for relieving stress?' (Lucy, Q. 19)
% summarization problem => incorrect retrieval + correct answer || vice versa
Our analysis identified CAIM's limitations regarding these tasks. For instance, summarization tasks showed different behaviors among the models. The GPT models surpass GLM by correctly summarizing movies or books, relying on their existing knowledge base rather than memory retrieval. In contrast, GLM did not rely on its pretrained knowledge. These findings lead to cases where the retrieval accuracy is rated 0 (= no relevant memory retrieved), but the response correctness is 1 (= correct answer), or vice versa. An example of incorrect retrieval but a correct answer is the response provided by GPT-4o: 'The world history-related game I recommended to you is likely Civilization [...]' to the question 'You once recommended a world history-related game to me. What was its name?' (Frank, A./Q. 14). In this case, the model successfully relied on its pretrained knowledge base to answer the probing question correctly, despite failing to retrieve the relevant memories. 
An example of correct retrieval and incorrect answer is GPT-3.5 turbo's response: '[...] you have not mentioned any specifics about your interest in outdoor activities. [...]' to the question 'I don't like outdoor activities, right?' (Jason, A./Q. 31). In this case, the model retrieved the correct memories but failed to utilize them to generate an accurate response. This aligns with the findings of Lu et al. \cite{lu2024insights}, who identified the 'Know but don't tell' phenomenon, in which LLMs accurately identify relevant information but fail to leverage this knowledge to generate an accurate response.

% exact list/recipe problem
Another identified limitation involves questions requiring exact lists or recipes. These questions challenge CAIM's ability to store and retrieve detailed information. For example, GLM responded to the question 'Do you remember the specific recipe?' (Ivy, Q. 53) with: '[...] I do remember the specific recipe, it was quite delicious. Would you like to share it with me now?' (Ivy, A. 53). This response demonstrates that CAIM's current memory storage and retrieval mechanisms are not designed to handle detailed information, which can result in difficulties in answering such questions correctly.

% time-based requests („first conversation“ vs. concrete dates)
\textbf{Time-based Requests:} Further findings include different ways of handling time-based requests across all models, identifying a potential area for improvement. Probing questions that included concrete dates, such as 'What topics did we talk about on April 29th?' (John Zhang, Q. 44), were mostly answered correctly due to CAIM's time-based relevance filtering, which prioritizes data with matching dates.
In contrast, questions without concrete dates, such as 'Do you remember what we talked about during our first conversation?' (Ivy, Q. 50), were answered incorrectly across all models, indicating a limited understanding of contextual references for relative time-based units, such as 'first conversation'.

% split information => contextually connect separate thoughts
\textbf{Split Information:} Another aspect identified in our analysis is CAIM's approach to split information. CAIM's memory mechanism divides relevant items into several inductive thoughts to ensure their conciseness. This approach leads to separate entries for related topics, which may hinder connecting contextually related thoughts, decreasing retrieval accuracy. 
For example, GPT-4o stored the following memories: 'suggested participating in programming-related communities' and 'System recommended maintaining enthusiasm for learning'. Despite both memories being relevant, GPT-4o could not successfully connect them to generate an accurate response to the question: 'I once consulted you on how to improve my programming efficiency. What advice did you give me?' (Roland, Q. 82). This identifies a limitation of CAIM's retrieval mechanism in connecting separate but contextually related pieces of information, pointing to a potential area for further improvement.

% Summary
The findings highlight the effectiveness, robustness, and context awareness of the Post-Thinking and Memory Retrieval modules. They demonstrate that CAIM effectively stores and recalls contextually relevant memories despite challenging conditions, such as rhetorical instead of fact-based questions. This highlights CAIM's ability to effectively address the outlined challenges, including the lack of LTM in LLMs and the need for more accurate and contextually relevant responses. Its handling of questions without relevant data demonstrates CAIM's robust approach addressing challenge 2. By correctly retrieving no memories when none are available, CAIM effectively avoids generating false positives, highlighting the strength of its filtering and retrieval mechanisms.

\subsection{Discussion}
% Summary Findings
Our findings demonstrate that CAIM successfully addresses the outlined challenges by ensuring high retrieval accuracy, leading to improved response correctness across all models. This supports CAIM's potential to enhance long-term interactions with LLMs.

% discussion 1: depth-level of information (efficiency vs. details)
However, our results also reveal CAIM's limitations in answering detailed queries, such as summarization tasks, recalling specific recipes, or providing an exact list of recommended items. These limitations occur because CAIM is instructed to store key pieces of interactions rather than every detail. This design aims to prevent the LTM from growing unnecessarily, which is consistent with earlier work that discusses the importance of concise memory structures to avoid overloading the system \cite{liu2021reuse, liu2023tim}. It minimizes the data retrieved, avoiding the risk of exceeding the context window and ensuring conciseness and that unnecessary details do not overload the system, which is a common challenge in memory-augmented systems \cite{yen2024memolet, zhong2024memorybank, liu2023tim, wang2023scm}

For such tasks that require a high level of detail, CAIM responds in a generalized manner, for example: 'I recommended to try some relaxing activities such as listening to music, watching movies, reading books or doing yoga' (Lucy, A. 19, GLM) or stores key events without details, such as 'Emily went to see the movie Taxi Driver' (Emily, A. 5, GPT-4o). While CAIM prioritizes storing key events, this approach may struggle with queries that require a high level of detail. This limitation highlights a trade-off within CAIM's architecture between conciseness and detail. We believe that the required level of detail in stored data depends on the considered application, and finding the right balance is crucial to optimize both memory management and user satisfaction. In addition, a more flexible memory mechanism that can adapt to various levels of detail based on the context could be an interesting approach to optimize for tasks such as recalling recipes.

% discussion 2: performance vs. word count
Furthermore, the results presented in Table \ref{tab:word_count} reveal the differences between models and suggest potential areas of improvement for CAIM. The findings indicate that balancing memory storage is essential, as GPT-3.5 turbo shows that storing more words does not necessarily lead to better results. More irrelevant data can reduce overall performance \cite{wang2023scm}, emphasizing the need to enhance filtering mechanisms to further minimize irrelevant data to improve CAIM. Additionally, GPT-4o demonstrates that a higher number of entries and more concise inductive thoughts lead to more efficient memory storage and retrieval. This also indicates that the entry type (concise thought; sentence) is essential for better performance. However, as mentioned before, the design of CAIM also highlights a limitation for split data. While concise thoughts improve efficiency, the underlying model may struggle to reconnect separate but contextually related pieces of information. Therefore, CAIM should prioritize quality, focusing on relevant, concise thoughts over quantity, such as detailed sentences, while future efforts should address its ability to link related entries to ensure coherent and contextually accurate responses. This challenge is further amplified by the generative variability of LLMs, which can produce diverse outputs even for the same input \cite{weisz2024design}.

\section{Conclusion}
In this work, we proposed CAIM, a holistic memory mechanism that enhances long-term interactions with LLMs by incorporating cognitive AI principles. CAIM consists of three modules that enable the storage and retrieval of user-specific information, support contextual and time-based relevance filtering, and ensure contextual coherence to successfully address the outlined challenges of existing work and LLMs. Furthermore, CAIM integrates key concepts of cognitive AI, such as thoughts and decision-making, to ensure context awareness and a more natural, human-like interaction.

As our results show, CAIM serves as an efficient framework for improving memory capabilities, enabling models to store and retrieve contextually relevant information. Our experiment highlights the advantages of extending LLMs with CAIM, demonstrating its ability to surpass the shortcomings of existing approaches. Our findings also reveal areas for improvement, such as the use of relative time-based units or split information, which will be addressed and integrated into CAIM in future work. Additionally, CAIM's design prioritizes conciseness in memory storage to avoid the risk of exceeding the context window and overwhelming the model. This can lead to challenges when answering detailed queries and meeting corresponding user expectations.

\section{Future Work}
As in recent research in cognitive AI, integrating a cognitive architecture can create AI systems that are more robust, efficient, and adaptable. Remembering and retrieving information lay the foundation of a cognitive architecture \cite{sun2024can, niu2024large} and enable contextually aware interactions. Therefore, we believe that CAIM yields a promising and extendable approach toward holistic memory modeling for LLMs that enables personalized context-aware interactions.
To further improve CAIM, we plan to address the limitations identified in our findings, focusing on mechanisms that balance detail and efficiency in memory storage and retrieval, mainly for handling detailed queries. Additionally, we aim to face the limitation of CAIM's memory, which is currently incapable of learning from queries. To overcome this limitation, we plan to integrate additional cognitive AI principles, such as discovery, which involves implicit pattern recognition to find connections within the data and deepen the understanding of the user. 
Finally, while the experiment shows CAIM's effectiveness on synthetic data, we intend to further validate our approach by integrating CAIM into applications and testing it in real-world scenarios with users, evaluating its impact on user experience and its potential to enhance personalization.

\bibliographystyle{ACM-Reference-Format}
\bibliography{sample-base}

\appendix
\section{Prompt List}

\subsection{Memory Controller}
\label{sec:Memory_Controller}

\textbf{Prompt Information necessary:} "Given a user command, determine whether answering the request requires recalling specific preferences or details shared earlier in the conversation that directly impact the response, in addition to understanding the current conversation context. Choose 'A' if previous information (e.g., preferences, likes, restrictions, recommendations) is crucial for providing a relevant or personalized response. Choose 'B' if the current request can be fully addressed without referring to prior context or preferences. Do not explain your choice, just choose 'A' or 'B'."

\textbf{Prompt Retrieval and Conversation:} "Given a user command, determine whether answering the request requires both recalling specific information or preferences shared in past sessions (historical memory) and understanding details or context from the current conversation (conversation context). Choose 'A' if both memory retrieval and the current conversation context are required to fully address the request. Choose 'B' if only one of these is sufficient. Do not explain your choice, just choose 'A' or 'B'."

\textbf{Prompt Retrieval:} "Given a user command, determine whether executing the command requires knowledge of historical or previous information about the user (such as data or context from earlier interactions) or whether it only requires understanding of the ongoing conversation content (such as recent questions and answers within the current session). Answer 'A' if it requires historical or previously stored information that is not part of the current conversation, or 'B' if it only requires information from the current conversation, including questions or context from the current session. Do not explain your choice, just choose 'A' or 'B'."

\subsection{Memory Retrieval}
\label{sec:Memory_Retrieval}
% Ontology
\textbf{Prompt Ontology Expansion:} "Given an ontology and the user input, analyze whether the existing ontology can adequately categorize and describe the user input with two or three words. If the user input is not adequately represented by the current ontology, expand the ontology by adding new categories, subcategories, or tags to properly capture them. Ensure that all new categories, subcategories, and words consist of only one word and are not already listed anywhere in the ontology. After making the necessary expansions, only return the updated ontology in JSON format, without any further information and without including the markdown marks '```json' and '```'. If no expansion of the ontology is necessary or the user command lacks context, just return OK, without any further information.  
The ontology: \{\},  
The user input: \{\}"

\textbf{Prompt Select Tags:} "Given a user command and an ontology, select two or three single-word tags from the ontology, that summarize the main topic or intent of the message the best. The tags should be relevant and concise, in a comma-separated list, and without explaining the information. Do not generate new tags, only select and use the single-word tags from the ontology. Respond following to this format: tag1,tag2,tag3. 
The ontology: \{\}"

\textbf{Prompt Relevance of Memories:} "Given a user command and a list of items, determine which items are relevant based on the user command to respond contextually correctly to the user command, even if it means contradicting the user's statement, return them without explaining the information. If none is relevant just return ''. 
The list of items: \{\}"

\subsection{Post Thinking}
\label{sec:Post_Thinking}

\textbf{Prompt Key Events:} "Analyze the following input and identify key pieces of personal information, interests, plans, habits, personal details, and recommendations that are important for future reference. List each detail in its simplest and most specific form. For each extracted detail, summarize it as a short thought, specifying the context and subject, and make sure to list separate pieces of information as individual entries. Use the exact timestamp that is attached. Ignore generic greetings, casual responses, or redundant statements. Return the output in a comma-separated format without line breaks, strictly following this structure: key piece;timestamp,key piece;timestamp."

\end{document}